\documentclass{article}

\usepackage{arxiv}

\usepackage[utf8]{inputenc} 
\usepackage[T1]{fontenc}    
\usepackage{hyperref}       
\usepackage{url}            
\usepackage{booktabs}       
\usepackage{amsfonts}       
\usepackage{nicefrac}       
\usepackage{microtype}      
\usepackage{lipsum}		
\usepackage{graphicx}
\usepackage{natbib}
\usepackage{doi}

\usepackage{subcaption}
\usepackage{bm}
\usepackage{multirow}
\usepackage[table]{xcolor}
\usepackage{colortbl}

\definecolor{color1}{RGB}{0, 255, 0}
\definecolor{color2}{RGB}{0, 0, 255}
\definecolor{color3}{RGB}{255, 0, 0}
\DeclareRobustCommand{\legendsquare}[1]{%
  \textcolor{#1}{\rule{1ex}{1ex}}%
}

\title{A deformation-based morphometry framework for disentangling Alzheimer's disease from normal aging using learned normal aging templates}

\author{ \href{https://orcid.org/0000-0003-4175-395X}{\includegraphics[scale=0.06]{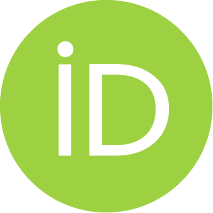}\hspace{1mm}Jingru Fu}\\
	Division of Biomedical Imaging\\
	KTH Royal Institute of Technology\\
	\texttt{jingruf@kth.se} \\
	\And
	\href{https://orcid.org/0000-0001-9522-4338}{\includegraphics[scale=0.06]{orcid.pdf}\hspace{1mm}Daniel Ferreira} \\
	Division of Clinical Geriatrics\\
	Karolinska Institute\\
        Facultad de Ciencias de la Salud\\
        Universidad Fernando Pessoa Canarias\\
\texttt{daniel.ferreira.padilla@ki.se} 
\\
\And
	\href{https://orcid.org/0000-0002-7750-1917}{\includegraphics[scale=0.06]{orcid.pdf}\hspace{1mm}Örjan Smedby} \\
	Division of Biomedical Imaging\\
	KTH Royal Institute of Technology\\
\texttt{orsme@kth.se} \\
 \And
	\href{https://orcid.org/0000-0001-5765-2964}{\includegraphics[scale=0.06]{orcid.pdf}\hspace{1mm}Rodrigo Moreno} \\
	Division of Biomedical Imaging\\
	KTH Royal Institute of Technology\\
	\texttt{rodmore@kth.se} \\
}

\renewcommand{\shorttitle}{\textit{arXiv} Template}

\hypersetup{
pdftitle={A template for the arxiv style},
pdfsubject={q-bio.NC, q-bio.QM},
pdfauthor={David S.~Hippocampus, Elias D.~Striatum},
pdfkeywords={First keyword, Second keyword, More},
}

\begin{document}
\maketitle

\begin{abstract}
Alzheimer's Disease (AD) and normal aging are both characterized by brain atrophy. The question of whether AD-related brain atrophy represents accelerated aging or a neurodegeneration process distinct from that in normal aging remains unresolved. Moreover, precisely disentangling AD-related brain atrophy from normal aging in a clinical context is complex. 
In this study, we propose a deformation-based morphometry framework to estimate normal aging and AD-specific atrophy patterns of subjects from morphological MRI scans. For this, we first leverage deep-learning-based methods to create age-dependent templates of cognitively normal (CN) subjects. These templates model the normal aging atrophy patterns in a CN population. Then, we use the learned diffeomorphic registration to estimate the one-year normal aging pattern at the voxel level.
In the second step, we register the testing image to the 60-year-old CN template. Finally, normal aging and AD-specific scores are estimated by measuring the alignment of this registration with the one-year normal aging pattern. 
The methodology was developed and evaluated on the OASIS3 dataset with 1,014 T1-weighted MRI scans, which is a unique dataset focused on preclinical cohorts. Of these, 326 scans were from CN subjects, and 688 scans were from individuals clinically diagnosed with AD at different stages of clinical severity defined by clinical dementia rating (CDR) scores.
The results show that ventricles predominantly follow an accelerated normal aging pattern in subjects with AD. In turn, hippocampi and amygdala regions were affected by both normal aging and AD-specific factors. Interestingly, hippocampi and amygdala regions showed more of an accelerated normal aging pattern for subjects during the early clinical stages of the disease, while the AD-specific score increases in later clinical stages.
Our code is freely available at \url{https://github.com/Fjr9516/DBM_with_DL}.
\end{abstract}


\section{Introduction}

\begin{figure}
\centering
\includegraphics[width=1\textwidth]{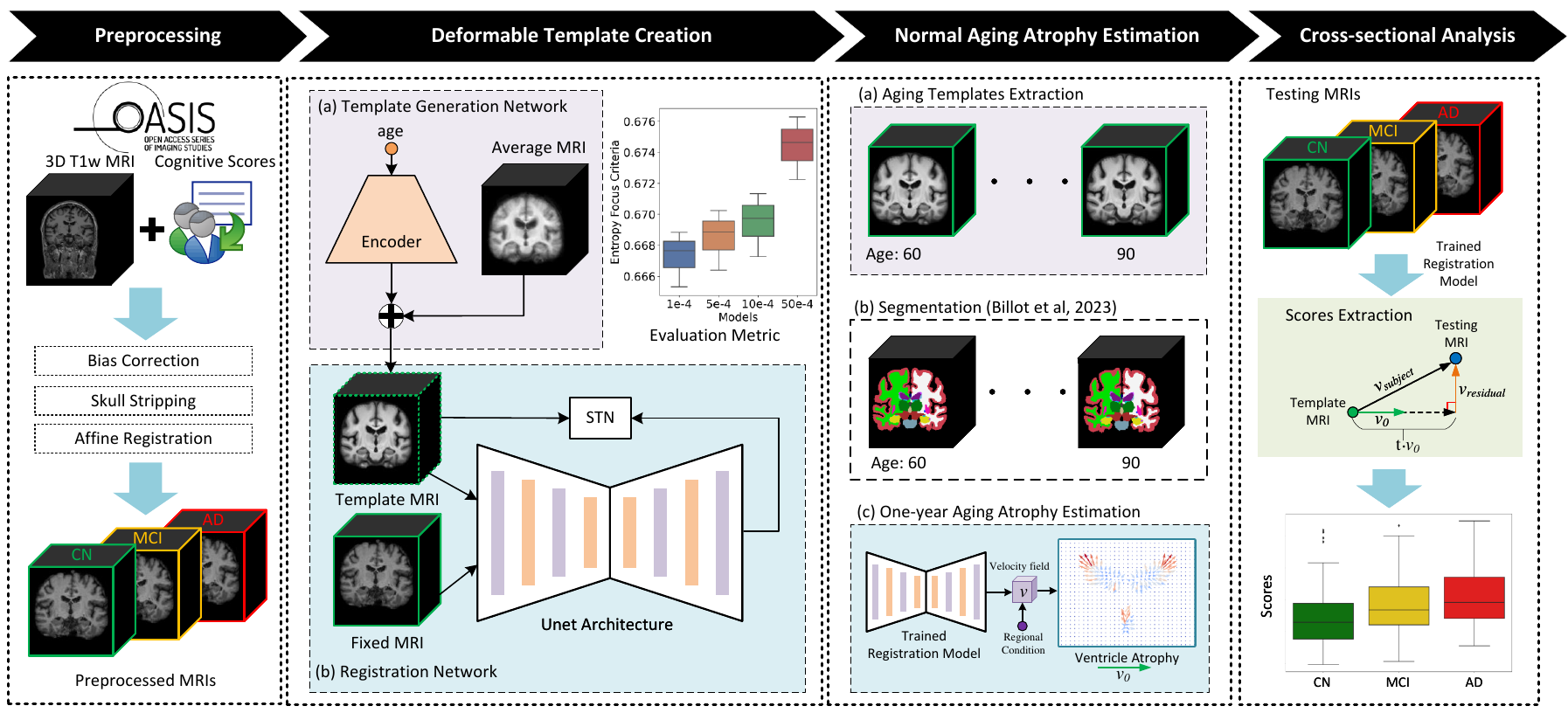}
\caption{Schematic of the proposed pipeline. 1) Preprocessing: OASIS-3 dataset was employed in this study, 3D T1w MRI scans and the corresponding subject-level clinical diagnosis were collected (i.e., CDR scores were used in this study), scans were processed following standard protocols and partitioned into cognitively normal (CN), mild cognitive impairment (MCI), and AD groups. Note that the three groups illustrated in this figure represent a simplified progression of AD; 2) A deep learning-based template creation method was employed to generate age-specific CN templates.
3) CN templates from 60 to 90 years old were generated and segmented. Deep learning-based registration was used to estimate the stationary velocity field (SVF) that describes the expected one-year healthy atrophy for the range between 60 and 90 years old. Segmentation masks were provided to obtain regional estimations of normal aging atrophy.
4) Deformation field between a given image and the 60-year-old template was decomposed into its aging and AD-specific components by comparing each vector with the one-year healthy atrophy vector field. These components were used to estimate aging score (AS) and AD-specific score (ADS). These scores were used for group comparisons.
}
\label{fig:overall}
\end{figure}

Alzheimer's disease (AD) is a serious public health concern as the aging population continues to grow, with an increasing number of cases each year. Currently, more than 55 million people worldwide suffer from dementia, with nearly 10 million new cases every year, of which AD is the most common form and may account for 60–70$\%$ of cases, as stated in the World Health Organization fact sheet (2023) \cite{WHO}. AD is characterized by progressive cognitive decline and is linked to brain atrophy and changes in brain structure~\cite{coupe2019lifespan}. Detecting and monitoring AD progression is crucial for early diagnosis, disease management and treatment~\cite{marquez2019neuroimaging}. Neuroimaging techniques such as magnetic resonance imaging (MRI) can be used to visualize and quantify changes in brain structure, making it a useful tool for studying age-related diseases like AD \cite{liu2003longitudinal}. Particularly, structural MRI (sMRI) is listed by the National Institute on Aging and the Alzheimer’s Association as part of the routine clinical assessment criteria for patients suspected to have AD \cite{jack2018nia}. 

Studies have shown that normal aging and AD are intertwined, and some researchers have hypothesized that AD is characterized by an accelerated global aging process~\cite{fjell2009one, whitwell2008rates, raz2005regional}. However, describing normal aging precisely in a clinical context is complex, as aging affects every part of the body with specific mechanisms and rates. Thus, multiple theories of aging have been proposed, leading to the definition of surrogate age variables based on the quantification of biological changes \cite{vina2007theories, sivera2019model}. Morphological changes in the brain can be characterized through Deformation-Based Morphometry (DBM) \cite{ashburner1998identifying}. This involves estimating spatial transformations (i.e., deformations) by non-linear registration, which enables the \textit{regional} analysis of the brain morphology. DBM helps study the progression of morphological changes in the brain observed in time series of images to model and quantify AD \cite{scahill2002mapping}. However, the morphological changes affected by AD are not completely related to the disease, especially in the asymptomatic and prodromal stages, because the brain structure is also influenced by patient phenotype and clinical history \cite{lorenzi2015disentangling}. A primary risk factor for AD is aging, which leads to patterns of structural loss that overlap with pathological loss. As claimed in previous studies, AD can be viewed as a pathological condition concurrent with aging, characterized by specific biochemical and structural hallmarks. The ability to independently model normal aging and AD would offer a means to delineate a given anatomy as being composed of distinct and concurrent factors. Such decomposition holds immense value, not only for advancing the understanding of the disease but also for clinical applications, including early diagnosis and the development of drugs targeting atrophy specific to the pathology. 

This study proposes a new DBM framework using learned CN templates for disentangling normal aging and disease-specific effects in AD. The methodology is divided into four main steps, as shown in Figure \ref{fig:overall}. First, we collected T1-weighted (T1w) MRI scans from the publicly available OASIS3 dataset and applied standardized preprocessing protocols to obtain preprocessed images. Second, a cohort consisting only of healthy subjects (normal aging, cognitively unimpaired) was identified, and a deep learning model was trained to create deformable templates of normal aging. The learned templates' “goodness” was monitored by their sharpness, and we confirmed that they accurately represented the underlying brain atrophy in the real data. Third, the learned CN templates were extracted at varying ages and utilized to discern age-related brain atrophy in new subjects. This involved the extraction of one-year normal aging atrophy patterns encoded in stationary velocity fields (SVFs). To facilitate regional-level analysis, an advanced segmentation method was employed to generate precise segmentation masks for the learned templates. Finally, a cross-sectional analysis was conducted, introducing two orthogonal scores to quantify the influence of both normal aging and AD-specific factors on a separate cohort of AD individuals at different clinical stages of the disease. Clinical stages of the disease were based on the well-established Clinical Dementia Rating (CDR) total score, while all AD individuals demonstrated progression to AD dementia at some point during follow-up clinical evaluations. This comprehensive approach enhances our understanding of the intricate relationship between normal aging and AD-specific progression, shedding light on the distinct influences in different brain regions.

The contributions of this work can be summarized as follows:
\begin{itemize}
  \item A novel DBM framework was introduced, integrating a learning-based template creation method. This advancement streamlined the entire DBM process, from subject-level morphology estimation to template space normalization.
  \item Deformable templates representing healthy aging were provided at the image level, accurately capturing morphological changes associated with normal aging.
  \item The introduction of the segmentation masks in the framework enables the identification of specific regional hallmarks for both normal aging and AD.
  \item By disentangling normal aging effects from AD progression, we unveiled variations in affected regions and clinical stages related to AD.
\end{itemize}

\section{Related Work}
Brain age prediction models based on brain MRI have become popular in neuroscience \cite{franke2012brain, liem2017predicting, cole2017predicting, brusini2022mri}, with the negative difference between predicted and chronological age thought to serve as an important biomarker reflecting pathological processes in the brain. Convolutional Neural Network (CNN) models have been successfully applied in brain MRI-based age prediction \cite{herent2018brain, li2018brain, cole2017predicting}. Several studies have shown the relationship between accelerated brain aging and AD \cite{gaser2013brainage}. However, the interpretability of this \textit{global} brain age biomarker is limited, particularly for neurodegenerative diseases that predominantly affect specific brain regions, such as the hippocampus in AD \cite{rao2022hippocampus}.

Several studies have shown the utility of DBM in distinguishing AD from normal aging \cite{hadj2016longitudinal, lorenzi2015disentangling}. In the standard DBM workflow, the morphological evolution of each subject is quantified through non-linear registration, and individual biological variability is normalized within a study-specific template space for group comparisons. Creating a study-specific template is crucial to account for inter-subject and inter-cohort differences that significantly influence the analysis of local anatomical variations. In the existing DBM studies on the brain \cite{cardenas2007deformation, lorenzi2011mapping, sudmeyer2012longitudinal, lorenzi2015disentangling, hadj2016longitudinal, sivera2019model}, all those steps involve substantial computational demands due to the high dimensionality of brain MRI data. This includes optimizing non-linear registration for each subject pair and potentially employing computationally intensive parallel transport algorithms to align individual trajectories to the study-specific template space. 

In recent years, the emergence of learning-based registration methods has been driven by advancements in deep learning \cite{yang2017quicksilver, balakrishnan2019voxelmorph, hoffmann2021synthmorph, chen2022transmorph}. These methods offer several advantages over classical registration approaches, ranging from direct knowledge extraction from data to enhanced efficiency during testing. The evolution of learning-based registration techniques has also spurred advancements in deformable template creation methods \cite{cabezas2011review, iglesias2015multi, niethammer2011geodesic}, which traditionally rely on non-linear registration models. Contemporary template learning methods encompass both implicit techniques, which update templates through averages of warped scans of subjects \cite{avants2004geodesic, avants2010optimal}, and explicit unsupervised deep networks for estimation \cite{dalca2019learning}. The objective of both approaches is to minimize image dissimilarity between the transformed template and fixed image while ensuring a diffeomorphic transformation via regularization. However, the intricate inter-brain variability entails complex topological differences that pure diffeomorphic models fail to capture. This limitation leads to anatomical boundaries in estimated templates that appear unrealistic and ambiguous. As a result, downstream applications like atlas-based segmentation and subsequent morphometric analysis may be affected \cite{senjem2005comparison, thompson2000mathematical}. To address this challenge, \citep{9711216} introduced generative adversarial learning into template learning, resulting in anatomically plausible templates with well-defined boundaries that are optimal for subsequent analysis. 

Motivated by regional DBM and DL-based template creation methods, we jointly merge these two approaches to leverage the rapidity of DL registration and the automatic learning capabilities of template generation. Furthermore, considering the evolution of aging as a dynamic process, the adoption of a learning-based \textit{conditional} template creation method in DBM prevents the need to bring subject-level trajectories to a common template space. Simultaneously, the integration of learning-enhanced registration models can be directly obtained following template creation method training. This holistic approach enhances the accuracy and efficiency of the entire DBM pipeline.
 
\section{Methods and Materials}
\begin{table}[!ht]
\renewcommand{\arraystretch}{1.5}
    \begin{center}
    \caption{Summary of OASIS-3 dataset}
    \label{tab:dataset}
    \begin{tabular}{ccc}
    \toprule
    {} & \textbf{Number of subjects} & \textbf{Number of T1w scans}\\
    \midrule
    Collected & 1316 & 2681\\
    \addlinespace
    \begin{tabular}[c]{@{}c@{}}
        Excluded\\
        (\textit{Quarantined} QC/undefined CDR)
    \end{tabular} & 1 & 3\\
    \addlinespace
    \begin{tabular}[c]{@{}c@{}}
        Remained\\
        (CN/AD dataset/non-AD dementia)
    \end{tabular} & 
    \begin{tabular}[c]{@{}c@{}}
        1315\\
        (739/419/157)
    \end{tabular} &
    \begin{tabular}[c]{@{}c@{}}
        2678\\
        (1678/688/312)
    \end{tabular}\\
    \addlinespace
    \midrule
    \begin{tabular}[c]{@{}c@{}}
        \textbf{Included in this study}\\
        \textbf{(CN/AD dataset)}\\
        \textbf{CDR for AD progression dataset at MRI (0/0.5/1/2)}
    \end{tabular} &
    \begin{tabular}[c]{@{}c@{}}
        1158\\
        (739/419)\\
        (143/246/94/4)
    \end{tabular} &
    \begin{tabular}[c]{@{}c@{}}
        2366\\
        (1678/688)\\
        (276/307/101/4)
    \end{tabular}\\
    \bottomrule
    \end{tabular}
    \end{center}
\end{table}

\subsection{Dataset}
The Open Access Series of Imaging Studies - version 3 (OASIS-3) dataset \citep{lamontagne2019oasi} was used to train the deformable template generation model and conduct the evaluations. OASIS-3 is unique in that it focused on a pre-clinical cohort and followed their longitudinal progress. OASIS-3 is a compilation of data for 1,378 participants, including 755 cognitively normal adults and 622 individuals at various stages of cognitive decline ranging in age from 42-95 years. It contains over 2,000 magnetic resonance (MR) sessions and includes T1 weighted (T1w) MR scans, among other sequences. Many of the MR sessions are accompanied by volumetric segmentation files produced through FreeSurfer processing\footnote{\url{https://surfer.nmr.mgh.harvard.edu/}}. OASIS-3 is a longitudinal multimodal neuroimaging, clinical, cognitive, and biomarker dataset for normal aging and AD\footnote{\url{https://www.oasis-brains.org/##about}}. In this study, we used T1w scans to address our study aims.

\begin{figure}[t]
     \centering
     \begin{subfigure}[b]{0.3\textwidth}
         \centering
         \includegraphics[width=\textwidth]{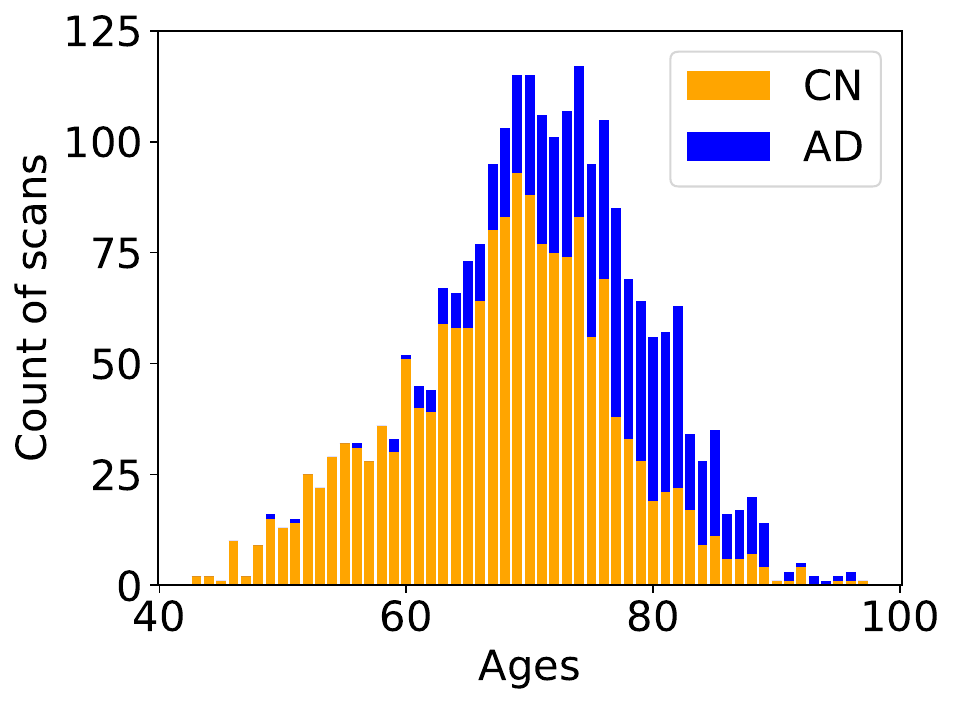}
         \caption{OASIS3 All scans (2366 images)}
         \label{fig:y equals x}
     \end{subfigure}
     \hfill
     \begin{subfigure}[b]{0.3\textwidth}
         \centering
         \includegraphics[width=\textwidth]{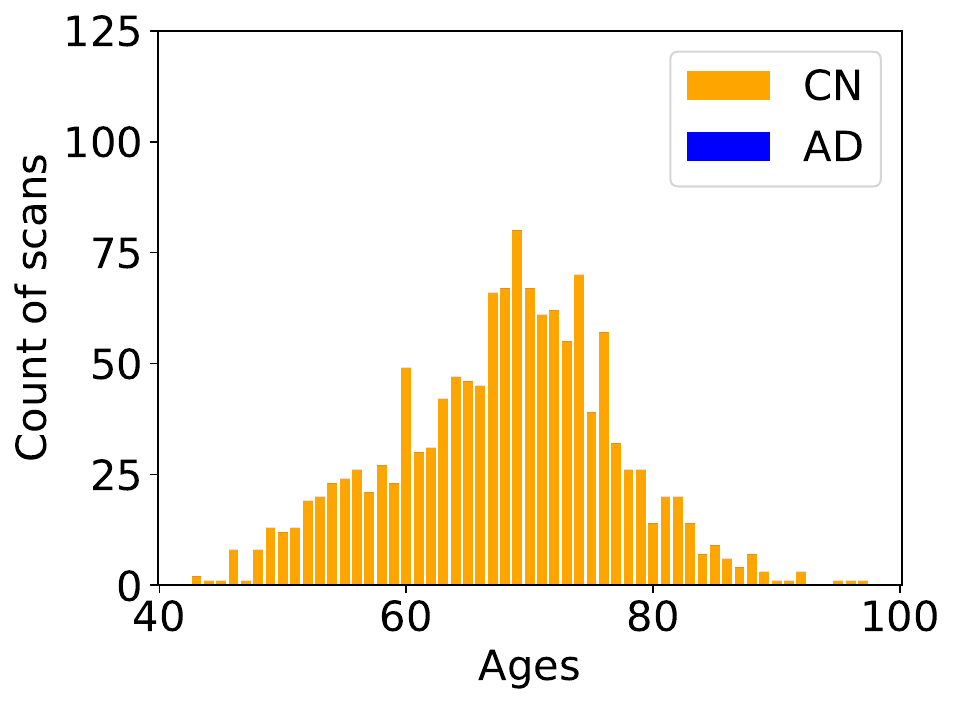}
         \caption{OASIS3 train scans (1352 images)}
         \label{fig:three sin x}
     \end{subfigure}
     \hfill
     \begin{subfigure}[b]{0.28\textwidth}
         \centering
         \includegraphics[width=\textwidth]{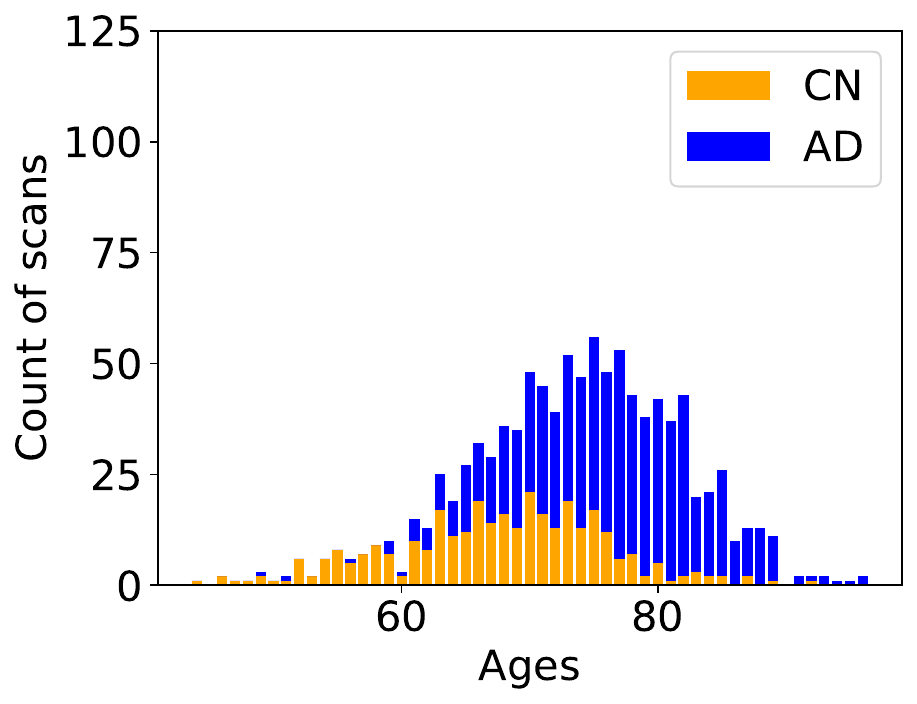}
         \caption{OASIS3 test scans (1014 images)}
         \label{fig:five over x}
     \end{subfigure}
        \caption{Histograms of sample size versus scan age for selected OASIS3 scans and the training and test sets constructed in this study. The oversampling process ensures that both training and test sets present a similar proportion over the available age range.}
        \label{fig:datasetPartition}
\end{figure}

\subsubsection{Data Preparation} 
We collected the FreeSurfer processed OASIS-3 dataset released in July 2022. A summary of this collection is shown in Table~\ref{tab:dataset}, in which the number of subjects and scans are counted separately. There were 2,681 successfully collected scans, in which two scans failed FreeSurfer Quality Control (QC) procedure (i.e., marked as \emph{Quarantined}) and one scan failed to define the CDR score, which is also an inclusion variable in our study. Therefore, these scans were discarded for the study. We also filtered the remaining scans according to the diagnoses provided in OASIS-3 and further removed non-AD dementia types to keep the focus on AD progression in our study. In this cohort, the clinical stage was defined by CDR following standards, as follows: a CDR of 0 corresponds to normal cognitive function, CDR = 0.5 indicates very mild cognitive impairment, CDR = 1 indicates mild dementia, CDR = 2 indicates moderate dementia, and CDR = 3 indicates severe dementia. According to OASIS-3's acquisition protocol, participants who reached CDR = 2 were no longer eligible for in-person assessments. For this reason, subjects with CDR = 2 are scarce in this dataset and CDR=3 are non-existent. After our selection steps and curation of data, 2,366 scans from 1,158 subjects were included in the study. We defined cognitively normal (CN) as subjects with CDR = 0 at all visits, and these were the scans used for the dataset of normal aging in our study. 1678 scans from 739 subjects complied with this requirement. In turn, the AD cohort was composed of individuals who progressed to clinical AD dementia at some point during follow-up visits but could have a CDR of 0 or >0 at the MRI scanning time. Notice that although some scans of the AD group can have CDR = 0, they are not mixed with the CN group since these images might already show some signs of AD progression that are not yet clinically signaled by the CDR test. Thus, we used 688 images from 419 subjects for the AD group with CDR scores from 0 to 2. Please see Table \ref{tab:dataset} for the breakdown of CDR scores in the AD progression dataset.

Our approach includes the use of templates for normal aging brains. For this, it is essential to ensure sufficient data coverage across different age groups. To achieve this, we employed an oversampling strategy that constructs the training and test sets in proportion to the age distribution of the CN datasets. Approximately 80$\%$ of the CN data was allocated to the training set, while the remaining 20$\%$ was assigned to the test set. Only the test set was used for the results reported in Sect. \ref{sec:results} below. Figure \ref{fig:datasetPartition} illustrates histograms that depict the partitioning of the dataset, with 1,352 images assigned to the training set and 326 images assigned to the test set. It is worth noting that the AD dataset, consisting of 688 images, is solely used for testing purposes.

\textbf{Neuroimaging Processing}:
We followed the protocol as previously outlined \cite{9711216} to prepare the suitable format data to train the deformable template generation model. Specifically, FreeSurfer processed OASIS-3 T1w data were obtained (i.e., \emph{norm.mgz}). FreeSurfer performs skull-stripping and bias field correction according to the FreeSurfer process flow\footnote{\url{https://surfer.nmr.mgh.harvard.edu/fswiki/ReconAllDevTable}}. Then, affine registration was applied to the image using the FreeSurfer \emph{mri\_vol2vol} command, utilizing Talairach space encoded in \emph{talairach.xfm} to Montreal Neurological Institute (MNI) 305 space. Then the segmentation masks for each image can be obtained by SynthSeg \citep{billot_synthseg_2023}, which is the state-of-the-art for brain image segmentation. To harmonize medical data for the deep learning model training, we rescaled the intensity of images to the range [0,1]. Finally, we cropped the input scan in the size of [208, 176, 160] to speed up training for this 3D problem and reduce computational demand. The size was calculated to preserve most foreground information using 200 randomly selected images from the dataset. 

\subsection{Deep learning-based deformable templates construction}
To capture population-level knowledge and account for individual variability, it is crucial to construct population-specific templates. In the context of deformable templates, deformable image registration is employed to create templates that are unbiased and barycentric, minimizing the average geodesic distance to each individual subject in the population \cite{avants2004geodesic, joshi2004unbiased, lorenzen2006multi, blaiotta2018generative, dey2021generative}. Traditional template creation methods are often time-consuming, particularly for obtaining high-resolution anatomical templates \cite{joshi2004unbiased, lorenzen2006multi}. However, recent advancements in deep learning have alleviated computational demands during the template estimation process \cite{dalca2019learning, dey2021generative}.

In this work, we selected AtlasGAN, a state-of-the-art method known for preserving distinguishable anatomical boundaries. AtlasGAN enables the estimation of realistic anatomy by utilizing generative adversarial learning to produce optimal conditional templates. The Deformable Template Creation section of Fig. \ref{fig:overall} shows a scheme of the method, which consists of three sub-networks: template generation,  registration and discriminator sub-networks.
The first two sub-networks are responsible for generating the conditional template and deforming it to match a fixed image, which is then evaluated by the discriminator sub-network. For more detailed information, please refer to the original paper \cite{dey2021generative}.

\textbf{Hyperparameter Selection}:
The generative adversarial training of AtlasGAN involves the joint optimization of two networks: one generator and one discriminator. The overall generator loss $L_{G}$ and discriminator loss $L_{D}$ can be summarized as follows:
\begin{equation}
L_{G}=L_{LNCC}+L_{Reg}(\phi)+\lambda_{GAN}L_{GAN},
\end{equation}
\begin{equation}
L_{D}=L_{GAN}+\lambda_{gp}L_{GP}.
\end{equation}
$L_{LNCC}$ represents the squared localized normalized cross-correlation, which ensures local standardization of the image intensity. $L_{Reg}(\phi)$ serves as the deformation regularization penalty, which ensures the smoothness and centrality of the displacement. In particular, it comprises three terms:
\begin{equation}
L_{Reg}(\phi) = \lambda_1{\|\bar{u}\|}_2^2 + \lambda_2\displaystyle\sum_{p\in\Omega}{\|\nabla u(p)\|}_2^2 + \lambda_3\displaystyle\sum_{p\in\Omega}{\|u(p)\|}_2^2    \label{eq:reg}
\end{equation}
where $p$ represents the voxels in the image space $\Omega$, $u$ indicates the spatial displacement that satisfies $\phi = Id + u$, and $\bar{u}$ is the moving average of $u$ over a window of $100$ updates, given by $\bar{u} = \frac{1}{n}\sum_{p\in\Omega}u(p)$. The first term of Eq. \ref{eq:reg} encourages small spatial displacements throughout the dataset, whereas the second and third terms aim to yield smooth and small individual deformations. $L_{GAN}$ represents the least-squares GAN loss, which is an adversarial term that ensures realism for the moving templates \cite{mao2017least}. For the discriminator, in addition to $L_{GAN}$, the gradient penalty $L_{GP}$ \cite{mescheder2018training} is incorporated to enhance the stability of the GAN training, which is is defined as $L_{GP} = \frac{1}{2}\mathbb{E}_{x\sim P_{real}}[\|\nabla D(x)\|]_2^2$ 
with $P_{real}$ being the distribution of the real images, and $\nabla D(x)$ is the gradient of the discriminator for a specific input image $x$. For more details, please refer to \citet{dey2021generative, dalca2019learning}. Following \cite{dey2021generative}, we set  $\lambda_{GAN}=0.1, \lambda_{1}=1, \lambda_{2}=1$, and $\lambda_{3}=0.01$. Finally, the optimal gradient penalty parameter $\lambda_{gp}$ depends on the training dataset. Thus, we trained four models using $\lambda_{gp}=[1e^{-4},5e^{-4},10e^{-4},50e^{-4}]$ to select the best performing hyperparameter. 

\textbf{Evaluation Metric}: To quantify the sharpness of the template, we used the entropy focus criterion (EFC) \cite{atkinson1997automatic}, which has been extensively used in previous studies \cite{esteban2017mriqc, joshi2004unbiased, liu2015low, rajashekar2020high, 9711216}. Specifically, the EFC, $E$ is defined as:
\begin{equation}
	E = - \sum _{i=1}^{N} \frac{B_{i}}{B_{max}} \ln\left[\frac{B_{i}}{B_{max}}\right],
\end{equation}
with $N$ being the number of image voxels, and $B_{i}$ the value of the i-th image voxel. The largest possible voxel brightness would be obtained if all the energy in the image were in one voxel, given by:
\begin{equation}
	B_{max} = \sqrt{\sum _{i=1}^{N}B_{i}^2}
\end{equation}
In this scheme, $E=0$ is achieved when all the image energy is located in one voxel and the remaining voxels are black, while the maximum entropy is achieved when the image is uniformly gray. In other words, the sharper the image, the smaller the EFC value.
We also applied a mask to the EFC to remove the background effect and normalized the EFC value to the range of [0, 1].

\subsection{Normal aging atrophy estimation through diffeomorphic registration} 
The estimation of normal aging atrophy was performed in three steps: 1) normal aging template estimation, 2) template segmentation, and 3) one-year normal aging atrophy estimation, as illustrated in the subsection “Normal Aging Atrophy Estimation" of Fig. \ref{fig:overall}. 

First, sharp and age-specific CN templates were extracted with the template generation block described in the previous subsection. These templates serve as barycentric representations at different ages. Next, the segmentation masks for each template are obtained using SynthSeg \cite{billot_synthseg_2023}. We used this method since it has been reported to be faster and more accurate than standard methods, such as FreeSurfer. To represent brain morphological changes, diffeomorphic non-linear registration is utilized \cite{hadj2016longitudinal, lorenzi2015disentangling, sivera2019model}. Diffeomorphic registration aims to learn a smooth, differentiable, and invertible transformation to avoid tissue folding in the registered biological image pairs (e.g., tissues should not fold or disappear with age \cite{fu2023fast}). The transformation, referred to as the diffeomorphic deformation field, is obtained by solving an ordinary differential equation (ODE) parameterized by a stationary velocity field (SVF) $\bm v$:
\begin{equation}
	\frac{d\bm\phi^{(t)}}{d t} = \bm v(\bm\phi^{(t)})
\end{equation}
where $\bm\phi^{(0)}$ is initialized with an identity transform. The final deformation field $\bm\phi^{(1)}$ is obtained by integrating over unit time as follows:
\begin{equation}
 \bm\phi=\bm\phi^{(1)} = \displaystyle \int_0^1  \bm v(\bm\phi^{(t)}) {d t}. \label{eq:intlayer}
\end{equation}

The SVF encodes the anatomical changes between the registration pair. Thus, the one-year normal aging atrophy can be estimated by dividing the SVF that represents the registration between two CN templates by their age gap. In this study, we used the SVF between the 60-year- and 90-year-old CN templates divided by 30 as the one-year normal aging atrophy map. Various studies have shown that different brain structures age differently \cite{walhovd2005effects, raz2006differential, azevedo2019contribution}. Thus, it is possible to perform regional analyses by using segmentation masks to derive regional normal aging atrophy maps. 

\begin{figure}[t]
     \centering
     \begin{subfigure}[b]{0.48\textwidth}
         \centering
         \includegraphics[width=\textwidth]{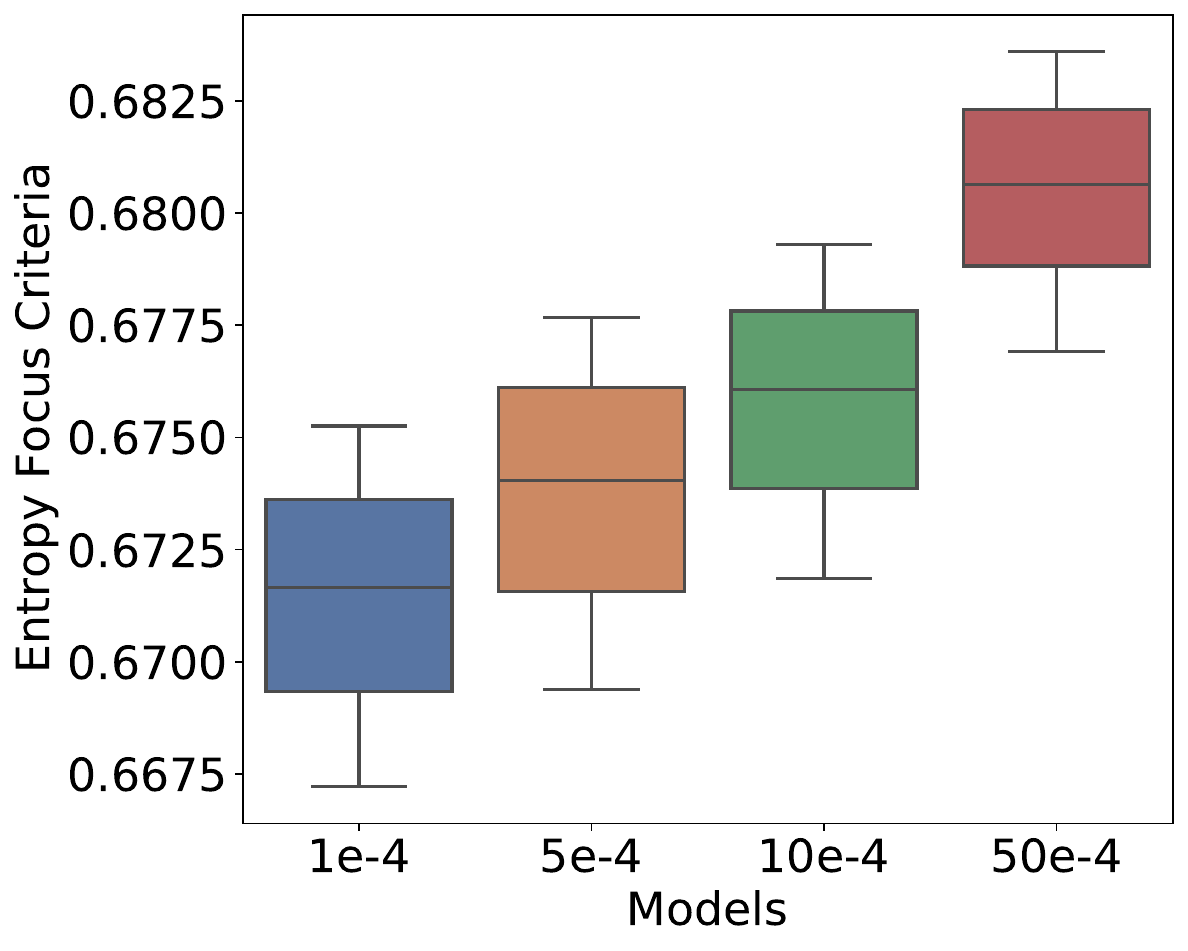}
         \caption{Models comparisons}
         \label{fig:EFCs}
     \end{subfigure}
     \hfill
     \begin{subfigure}[b]{0.49\textwidth}
         \centering
         \includegraphics[width=\textwidth]{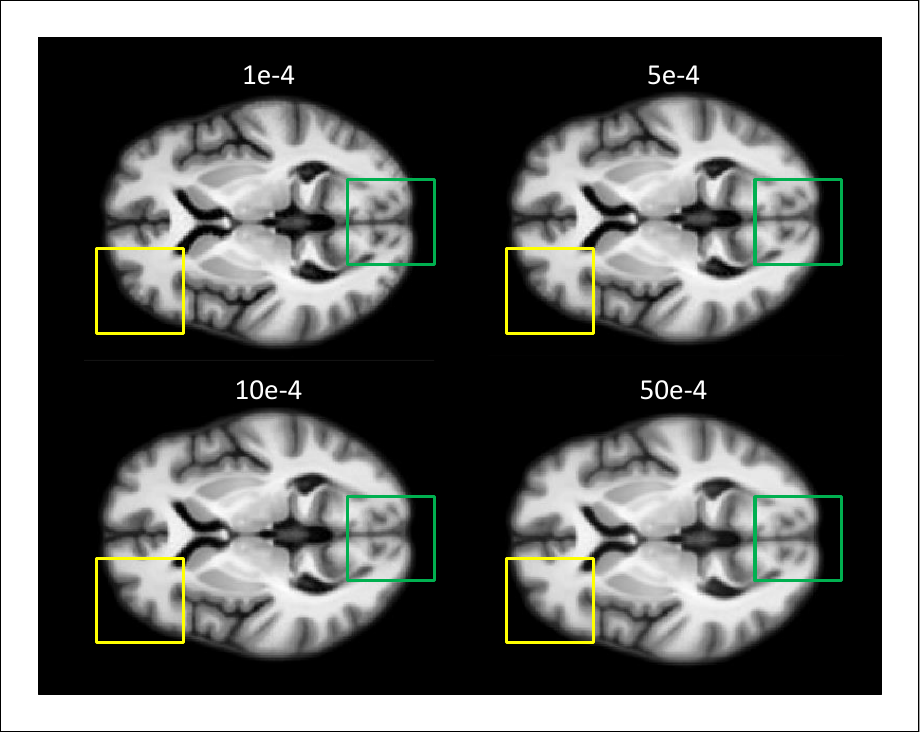}
         \caption{CN templates extracted from different models}
         \label{fig:sharp}
     \end{subfigure}
        \caption{(a) Boxplots for entropy focus criterion (EFC) scores on learned deformable templates using different hyperparameters for $\lambda_{gp}$. (b) Healthy templates at age 60 extracted from different models confirmed the sharpness could be well expressed using EFC. Two emphasized regions are boxed, indicating detailed structures represented in smaller EFC.}
\label{fig:sharp_segmentation}
\end{figure}

\subsection{Extraction of aging and AD-specific scores} 
This paper introduces the aging (AS) and AD-specific scores (ADS), as shown in the section “Cross-sectional Analysis" of Fig. \ref{fig:overall}. 
For this, we build upon the assumption by \cite{lorenzi2015disentangling} that states that normal aging and disease-specific components of a specific subject can be disentangled by orthogonally projecting the subject-to-template SVF $\bm v_{subject}$ onto the one-year normal aging SVF $\bm v_{0}$. In particular, the AS and ADS are computed as follows:
\begin{equation}
  AS = \frac{<\bm v_{subject}, \bm v_{0}>}{||\bm v_{0}||^2}   
\end{equation}
\begin{equation}
  ADS = ||\bm v_{subject} - AS \cdot \bm v_0||, 
\end{equation}
where $<\cdot,\cdot>$ is the inner product of vectors. These two scores are computed voxel-wise. AS measures the alignment of specific aging of the subject with 
the aging process in a CN population.

Negative and positive values of AS indicate that the subject is aging at a lower or higher pace than the CN population, respectively. 
Notice that the AS only captures variability within normal aging. That means that a specific subject might look very old (or very young) with respect to a population and still be cognitively normal. In turn, ADS aims to capture patterns that are not present in the CN population. Thus, ADS could potentially be more discriminative of patients and controls than AS.

The clinical onset of AD is usually around 65 years of age, but the disease process is known to start a decade or more before the clinical onset of dementia \cite{rabinovici2016testing}. Thus, we chose in this study the 60-year-old reference healthy template for performing the subject-to-template diffeomorphic registration.

\textbf{Outlier Rejection:} The calculation of the scores involves averaging voxel-by-voxel values, making it sensitive to outliers caused by the inter-subject difference and registration algorithm. The registration algorithm balances optimal alignment and diffeomorphism, resulting in inherent errors. Additionally, the scores for each voxel are directly influenced by the magnitude of the unit year healthy atrophy SVF $\bm v_{0}$. Notably, when $\bm v_{0}$ is small, it can lead to disproportionately large score values, potentially distorting the final score calculation. To address this, we implemented a thresholding strategy based on the norm of $\bm v_{0}$ to reject possible outliers for the final calculation. Our experiments show that this strategy is appropriate to deal with this issue.

\section{Results}
\label{sec:results}
\subsection{Deep learning model can create sharp longitudinal healthy templates}
\textbf{Sharp templates:} Imaging templates with distinguishable boundaries among brain regions are advantageous for increasing the accuracy of downstream tasks such as registration or segmentation \cite{9711216}. To quantify the sharpness of the images, we applied the entropy focus criterion (EFC) to the learned templates \cite{atkinson1997automatic}. We extracted healthy templates at each integer age from 60 to 80 years old since this range is the most common in the AD population overall and is well represented in the dataset. According to the results shown in Fig. \ref{fig:EFCs}, the model trained with $\lambda_{gp}=1e^{-4}$ showed the smallest  EFC (i.e., sharpest) templates, so it was chosen to be used in the subsequent analyses. For visualization purposes, the emphasized regions from the template of 60 years old at different models also confirmed the sharpness value as shown in Fig. \ref{fig:sharp}.  
 
\begin{figure}
	\centering
	\includegraphics[width=\textwidth]{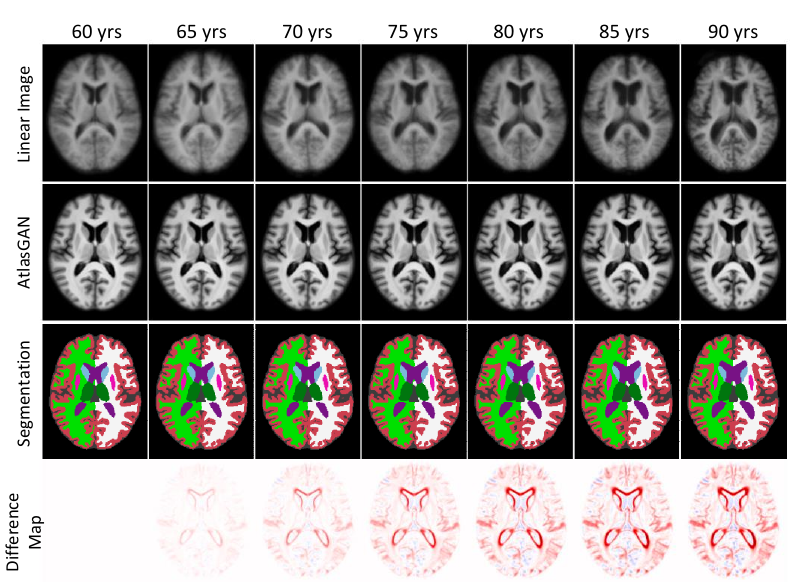}
	\caption[Caption for LOF]{Learned example templates with the corresponding segmentations from the axial plane at the 80th slice. \emph{First row}: The linear average images conditioned on different ages; \emph{Second row}: The learned templates conditioned on different ages; \emph{Third row}: The segmentation masks obtained by \cite{billot_synthseg_2023}. The segmentations are shown using the same color scheme used in  Freeview\footnotemark; \emph{Fourth row}: residual maps obtained by the corresponding column minus the first template.}
	\label{fig:templates}
\end{figure}
\footnotetext{\url{https://surfer.nmr.mgh.harvard.edu/fswiki/FsTutorial/AnatomicalROI/FreeSurferColorLUT}}

\begin{figure}[t]
     \centering
     \begin{subfigure}[b]{0.48\textwidth}
         \centering
         \includegraphics[width=\textwidth]{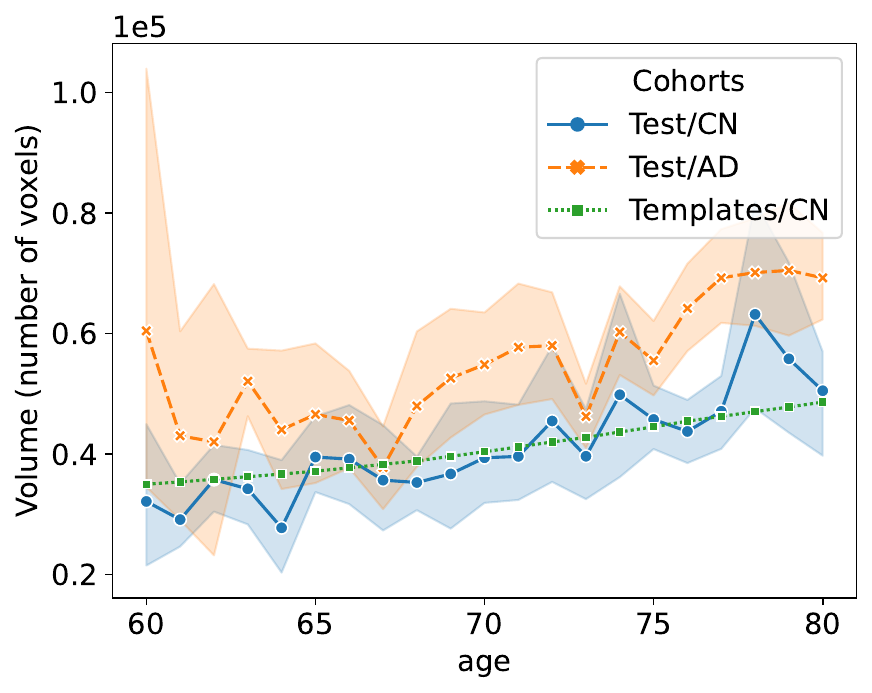}
         \caption{Ventricles}
         \label{fig:ventricles_syntheSeg}
     \end{subfigure}
     \hfill
     \begin{subfigure}[b]{0.48\textwidth}
         \centering
         \includegraphics[width=\textwidth]{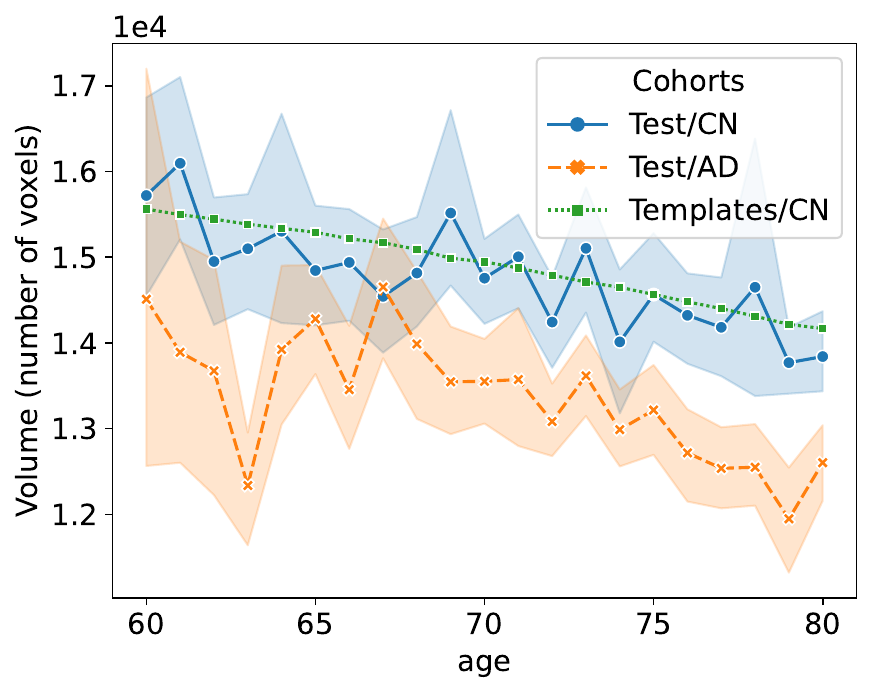}
         \caption{Hippocampi $\&$ Amygdala}
         \label{fig:hippocampus_syntheSeg}
     \end{subfigure}
        \caption{(a) Ventricles volumetric trend of learned templates and the real trends for AD and CN cohorts in the test sets, respectively. (b) Hippocampi $\&$ Amygdala trends of learned templates and the real trends for AD and CN cohorts in test sets, respectively. The standard deviations of AD and CN testing subjects are shown in light orange and blue, respectively.}
\label{fig:trends}
\end{figure}

\textbf{Comparison with real data:} The quality of synthetic templates is evaluated via the sharpness metric, but there is a need to evaluate the ability of learned templates to represent longitudinal changes from a morphological perspective. The learned templates are shown in Figure \ref{fig:templates}. As shown, the learned templates can provide sharp and clear boundaries compared to linear average scans. The first row was obtained by the average of randomly sampling 15 scans from the adjacent $\pm$ 2.5 years old range at each column. The second row is obtained by feeding the corresponding age value to the learned encoder block. The third row was the corresponding segmentation masks for each healthy template by using SynthSeg \cite{billot_synthseg_2023}. The fourth row was obtained by subtracting the first template, i.e., the template at age 60, from the image in each column. From this qualitative perspective, we could observe that learned templates successfully capture the morphological changes with age.

We evaluated the learned templates quantitatively by comparing the morphological differences represented by segmentation label counts with the underlying real data. We selected the ventricle volumetric trend together with hippocampi and amygdala regions in CN and AD cohorts \footnote{Segmentation labels: [4, 14, 15, 43] for ventricles and [17, 53, 18, 54] for Hippocampi $\&$ Amygdala}, and templates. As shown in Fig. \ref{fig:trends},  the brain volume progression associated with the model of normal aging entirely falls into the standard deviation of the underlined real data and can also be distinguishable from the AD testing cohort.  

\subsection{Aging and AD-specific scores applied to cognitively normal subjects}

\begin{figure}[t]
     \centering
     \begin{subfigure}[b]{\textwidth}
         \centering
         \includegraphics[width=\textwidth]{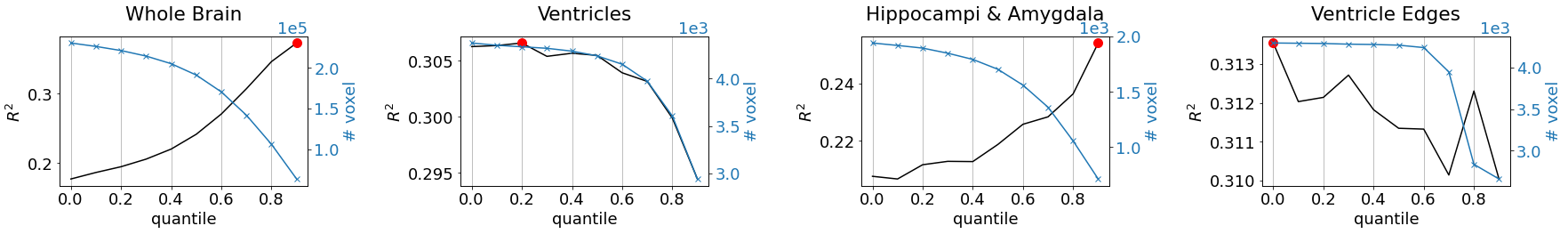}
         \caption{
         }
         \label{fig:R2_num}
     \end{subfigure}
     \hfill
     \begin{subfigure}[b]{\textwidth}
         \centering
         \includegraphics[width=\textwidth]{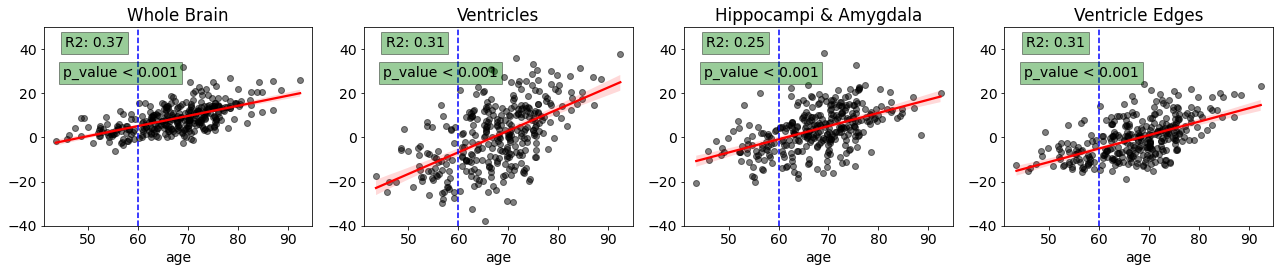}
         \caption{}
          \label{fig:markers_on_HC}
     \end{subfigure}
     \hfill
     \begin{subfigure}[b]{\textwidth}
         \centering
         \includegraphics[width=\textwidth]{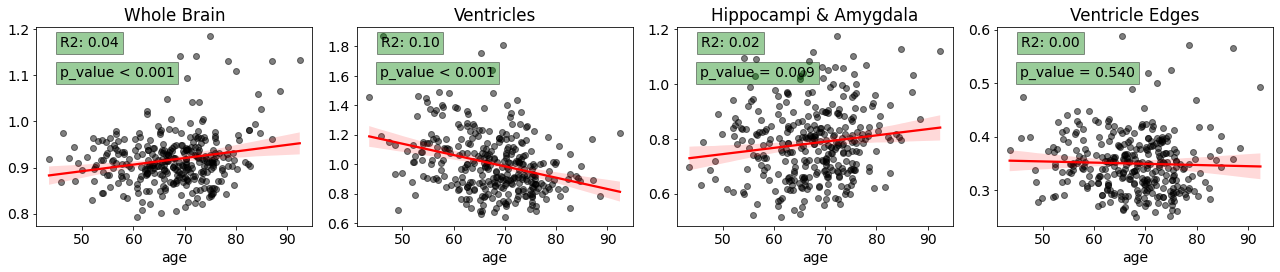}
         \caption{}
          \label{fig:ad_markers_on_HC}
     \end{subfigure}
        \caption{
        (a) $R^2$ and the number of preserved voxels after thresholding $\bm ||v_0||$ with different quantiles for each region. The highest $R^2$ is indicated by a red dot in the figure. (b) The relationship between AS and age per brain region on the CN test set. The reference age is indicated by the blue dashed line. (c) The relationship between ADS and age per brain region on the CN test set. The fitted lines, shown in red, are accompanied by 95$\%$ confidence interval (CI) shadow regions in both AS and ADS plots.  }
        \label{fig:R2_num_markers_on_HC}
\end{figure}

Once we obtain the CN templates, we can determine the most representative one-year normal aging atrophy $\bm v_0$ by extracting the SVF between two CN templates at different ages and dividing it by the age gap between the input image pair. 
Through experimentation, we found that the SVF between $T_{60}$ and $T_{90}$ provided better distinguishability for subsequent cross-sectional analysis. In our experiments, SVFs created from templates with a smaller age difference were less consistent. This could be attributed to the fact that SVFs can become less reliable when the morphological changes in the brain are more subtle. 

Since the AS can be affected by outliers, we used different quantile values ranging from 0 to 0.9 with an interval of 0.1 to threshold $\bm v_0$. We aimed to find the quantile value that optimizes $R^2$ value, which indicates the goodness of fit between the AS and age for the CN group. The test set consisted of scans from subjects with consistent CDR scores of 0 over scanning sessions. As mentioned before, a mask was applied in every calculation to mitigate the influence of the background.

We tested AS on four different brain regions: the whole brain, ventricles, hippocampi $\&$ amygdala, and the ventricle edge map. The ventricle edge map represents a region close to the edges of the ventricles and is defined as the difference between the ventricles segmentation of the 90-year-old and 60-year-old templates. The results are shown in Figure \ref{fig:R2_num}. As depicted, different quantile values yielded the best fit for different brain regions, indicating varying optimal quantile values for different brain regions. The number of preserved voxels at each quantile value is shown on the right y-axis, showing a sufficient number of preserved voxels after thresholding.

The optimal fitted results for AS are presented in Figure \ref{fig:markers_on_HC}, where the x-axis represents the chronological age and the y-axis shows the AS.
When morphological age is identical to morphological age, the fitted line should be equal to 0 at 60 and exhibit linear changes with age \cite{salat2004thinning, abe2008aging}. As shown, the ventricle region provides the best fit on the slope, indicating a closer alignment with chronological aging. However, the other three plots exhibit slower changes compared to chronological aging in cognitively normal individuals.
It is important to note that different brain regions exhibit distinct slopes in relation to the scores, indicating variations in aging rates across different brain regions. We also reported the corresponding fits for ADS in Figure \ref{fig:ad_markers_on_HC}, where we observed notably low $R^2$ scores for all four regions, indicating a weak dependence of ADS on age. However, the obtained low p-values for the whole brain and ventricles for ADS suggest the importance of accounting for age dependencies. 

\subsection{Aging and AD-specific scores can be used to assess the progression of AD}
\begin{figure}[t]
     \centering
         \centering
         \includegraphics[width=0.5\textwidth]{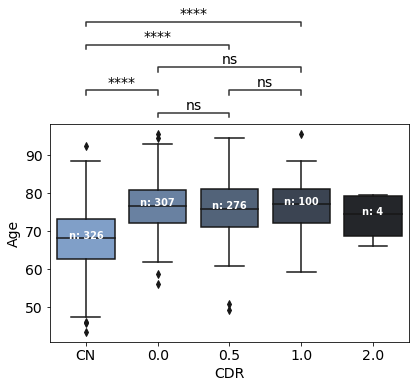}
        \caption{Chronological age distributions of the different cohorts (number of observations is shown in the middle of each boxplot as $n$). No significant difference was observed among the AD subgroups regarding chronological ages.}\label{fig:Age_dist}
\end{figure}

\begin{figure}[t]
     \centering
    \begin{subfigure}[b]{\textwidth}
         \centering
         \includegraphics[width=\textwidth]{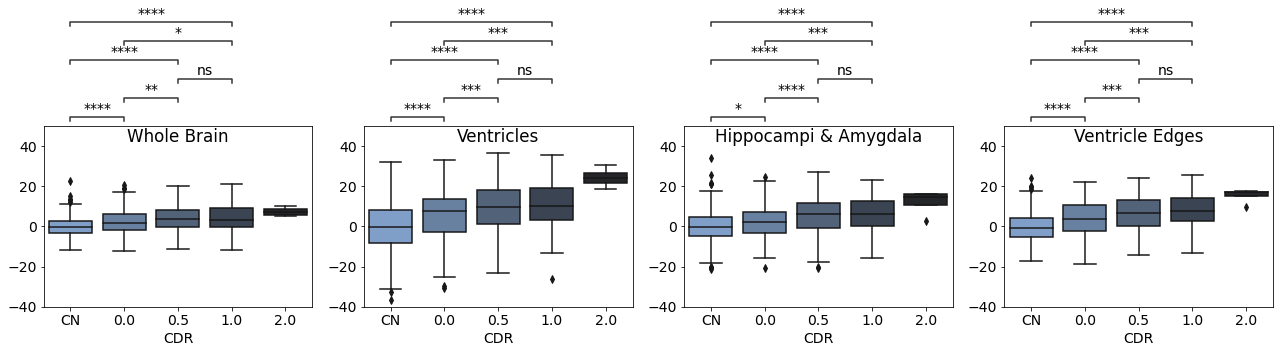}
         \caption{Aging score (AS) }
         \label{fig:markers_on_AD}
     \end{subfigure}
     \hfill
     \begin{subfigure}[b]{\textwidth}
         \centering
         \includegraphics[width=\textwidth]{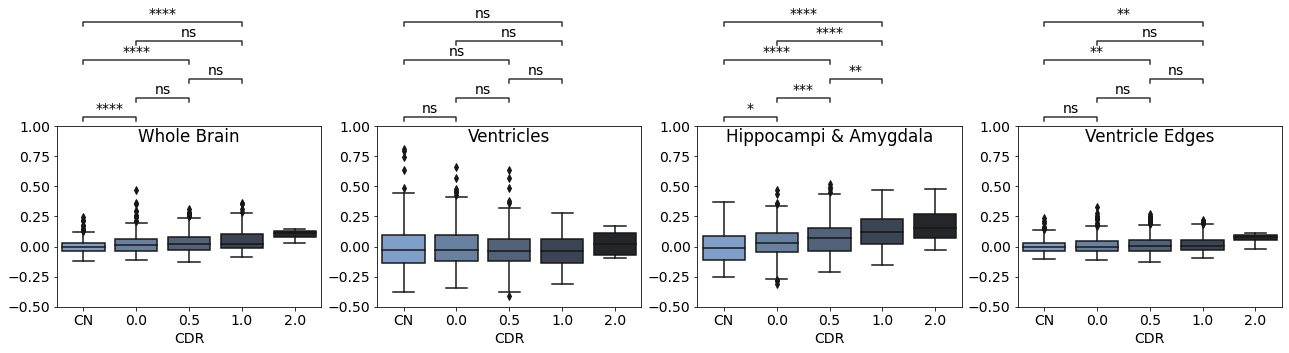}
         \caption{AD-specific score (ADS)}
          \label{fig:mag_on_AD}
     \end{subfigure}
        \caption{Distributions of the aging and AD-specific scores for different disease stages defined by CDR and cognitively normal subjects (CN).}
\label{fig:markers_mag_on_AD}
\end{figure}

To evaluate the proposed scores' ability to differentiate the progression of AD, the AD group was divided into four subgroups (i.e., CDR=0, CDR=0.5, CDR=1, and CDR=2) according to CDR scores at the time of MRI scanning. As mentioned before, all these subjects progressed to clinical AD dementia at follow-up evaluations. Independent t-tests were performed on each pair of distributions, and the Bonferroni method was used for multiple testing corrections \footnote{The p-value annotation legends in the Fig. \ref{fig:Age_dist} and \ref{fig:markers_mag_on_AD} are as follows: ns: p > 0.05;  *: 0.01 < p $\le$ 0.05;  **: 0.001 < p $\le$ 0.01; ***: 0.0001 < p $\le$ 0.001; ****: p $\le$ 0.0001. }. In the course of this experiment, a notable outlier emerged within the AD subgroup with CDR = 1. Upon visual inspection, it was determined that this outlier had failed to undergo proper affine registration in the FreeSurfer workflow. As a result, it was excluded from subsequent experiments.

Given the significant correlation of AS with chronological age, as illustrated in Figure \ref{fig:markers_on_HC}, we examined the distributions of chronological age among the CN cohort and the subgroups of AD individuals stratified into clinical stages based on CDR, as shown in Figure \ref{fig:Age_dist}. The results revealed no significant age differences between any pair of AD subgroups, while a noteworthy age-related difference was observed when comparing the CN group to the AD groups.

To remove the influence of age in the AS and ADS, a normalization procedure was employed. This involved the utilization of Analysis of Covariance (ANCOVA) for each score across levels of categorical disease groups, while statistically controlling for the age effect. The resulting adjusted scores were then harmonized on the same scales by subtracting the mean of the CN group for all four observed brain regions to facilitate more meaningful comparisons. As shown in Fig. \ref{fig:markers_on_AD}, there are increasing trends in the AS with respect to increasing disease stages defined by CDR scores in all four regions. This finding supports the assumption that AD is a factor for accelerated normal aging. 
Moreover, significant differences were evident in several AS comparisons. Specifically, there were significant differences for all CN vs. CDR = 0.0, and CDR = 0 vs. CDR = 0.5 adjacent AD dataset pairs. Regarding ADS (see Fig. \ref{fig:mag_on_AD}), it is worth noting that the hippocampus and amygdala regions exhibit the smallest p-values denoting the most statistical significance among the cohorts, which aligns with previous studies. The ADS shows significant differences in the hippocampus and amygdala and no significant differences in ventricle-related regions within AD subgroups.
Notably, when examining disease progression, the ADS can provide supplementary information for distinguishing AD progression, as it reveals a significant difference between the CDR=0.5 vs. CDR=1 adjacent AD pairs, which may not be readily discernible when considering only AS.
It is important to note that the observation for CDR = 2 is based on the limited availability of only four scans in the dataset. Thus, statistical testing was not performed for the CDR = 2 group.

Furthermore, in addition to performing independent t-tests, we reported Cohen's d effect sizes for AS and ADS. A conventional interpretation of effect sizes categorizes them as small (d = 0.2), medium (d = 0.5), and large (d = 0.8), as suggested by \citep{cohen2013statistical}. However, it is essential to note that these benchmarks should not be rigidly applied and must be considered in the context of specific research \cite{lakens2013calculating}. Normally, a Cohen's d of 0.5 signifies a difference equivalent to half a standard deviation. In light of this, we have further categorized effect sizes into more refined ranges based on Cohen's d absolute values, ranging from medium to very large \footnote{Medium effect size: [0.35, 0.65); large effect size: [0.65, 0.9); very large effect size: > 0.9}, and highlighted them with distinct colors in Table \ref{tab:effectsize}. 

\begin{table}[t]
\centering
\renewcommand{\arraystretch}{1.2}
\caption{Cohen’s d effect size measure for the statistical tests of Figure \ref{fig:markers_mag_on_AD}. The pairs match cognitively normal subjects (CN) and AD subjects with different CDR scores (0, 0.5, or 1). Note that the negative Cohen's d indicates that the mean of the first group is higher than that of the second group, while the value is related to the degree of effect size. \legendsquare{color1}: Medium effect size,
  \legendsquare{color2}: Large effect size,
  \legendsquare{color3}: Very large effect size.}
\begin{tabular}{cc|cccccc}
\hline
\multicolumn{2}{c|}{\multirow{2}{*}{\textbf{Regions}}} & \multicolumn{6}{c}{\textbf{Pairs}}                           \\ \cline{3-8} 
\multicolumn{2}{c|}{}                                  & (CN, 0) & (CN, 0.5) & (CN, 1) & (0, 0.5) & (0, 1) & (0.5, 1) \\ \hline
\multirow{2}{*}{Whole Brain}                 & AS      & \cellcolor{color1!20}-0.40   & \cellcolor{color2!20}-0.70     & \cellcolor{color2!20}-0.82   & -0.27    & \cellcolor{color1!20}-0.35  & -0.07    \\
                                             & ADS     & \cellcolor{color1!20}-0.36   & \cellcolor{color1!20}-0.48     & \cellcolor{color2!20}-0.75   & -0.10    & -0.29  & -0.20    \\ \hline
\multirow{2}{*}{Ventricles}                  & AS      & \cellcolor{color1!20}-0.43   & \cellcolor{color2!20}-0.74     & \cellcolor{color3!20}-0.90   & -0.32    & \cellcolor{color1!20}-0.47  & -0.15    \\
                                             & ADS     & -0.01   & 0.13      & 0.16    & 0.13     & 0.17   & 0.05    \\ \hline
\multirow{2}{*}{Hippocampi \& Amygdala}      & AS      & -0.22   & \cellcolor{color1!20}-0.61     & \cellcolor{color2!20}-0.68   & \cellcolor{color1!20}-0.41    & \cellcolor{color1!20}-0.49  & -0.06    \\
                                             & ADS     & -0.24    & \cellcolor{color1!20}-0.56     & \cellcolor{color3!20}-0.95   & -0.32    & \cellcolor{color2!20}-0.71  & \cellcolor{color1!20}-0.37    \\ \hline
\multirow{2}{*}{Ventricle Edges}             & AS      & \cellcolor{color1!20}-0.47   & \cellcolor{color2!20}-0.80     & \cellcolor{color3!20}-1.04   & -0.31    & \cellcolor{color1!20}-0.50  & -0.19    \\
                                             & ADS     & -0.15   & -0.26     & \cellcolor{color1!20}-0.38   & -0.09    & -0.18  & -0.09    \\ \hline
\end{tabular}\label{tab:effectsize}
\end{table}

\section{Discussion}
This study presents a framework to investigate whether AD accelerates normal aging or manifests as a distinct AD-specific process in terms of brain atrophy. Overall, the proposed framework aims to advance our understanding of the intricate interplay between normal aging and AD progression. By combining deep learning techniques, diffeomorphic registration, and biomarker analysis, we offer insights into the potential mechanisms underlying AD-related changes in brain structure. This study contributes to the broader field of neurodegenerative research by shedding light on the distinct components of brain alterations in AD and their relationship with the normal aging process. The proposed framework encompasses several steps that collectively contribute to our understanding of this complex relationship.

\subsection{Generation of age-specific templates}
Accurately describing the voxel-level changes associated with normal brain aging is a complex task, necessitating the consideration of both spatial (inter-subject variation) and temporal (intra-subject registration) dynamics. Many neuroimaging studies use standard templates (e.g., the classical MNI-152 template) or one of the images from the dataset to bring all images into the same space before further analyses. 
One of the issues of the former approach is that it is relatively blurred, compromising the accuracy of registrations. In turn, a specific subject might not be at the barycenter of the dataset, leading to biases. Also, both methodologies suffer from not being age-specific.

For this reason, we use a DL-based method to generate sharp, age-specific templates representing the CN subjects in the OASIS-3 dataset. In particular, we trained the neural network proposed by \cite{9711216} with our data. As sharpness is desirable to increase the accuracy of downstream processing (e.g. segmentation and registration), we test different hyperparameters to minimize the EFC (sharper images have lower EFC). As shown in Fig. \ref{fig:sharp_segmentation}, $\lambda_{gp}=1e^{-4}$ yields the best results.

However, it is worth noting that the pursuit of sharper images for the registration algorithm introduces certain limitations, as discussed in \cite{hadj2016longitudinal}. Specifically, a high-intensity gradient along boundaries can lead to larger deformations outside the target organ, in our case, the brain. This phenomenon might significantly impact deformation-dependent measurements. 
Fortunately, \cite{hadj2016longitudinal} has shown the efficacy of introducing masks to mitigate the influence of background artifacts. In our study, we also address this by introducing the segmentation mask, which helps counteract the impact of background errors.
 
As discussed in \cite{fu2023fast}, anatomical plausibility is usually an issue for DL-based image generation. To target this issue, state-of-the-art DL-based diffeomorphic registration methods integrate a non-DL-based module to integrate the SVFs \cite{hoffmann2021synthmorph}. This way, diffeomorphism can be guaranteed and, hence, the anatomical plausibility of the templates.
Moreover, the diffeomorphic registration module in the framework ensures that the learned templates accurately represent training samples proximal to the target age while preserving inter-subject topology. The continuous age condition further contributes to the templates' coherent evolution. As depicted in the first row of Figure \ref{fig:templates}, linear templates show an exaggerated variation of the asymmetric ventricular size with age (e.g., compare the 60-year-old and the 90-year-old linear templates). In turn, the learned templates successfully mitigate this bias, yielding a more symmetric evolution with age.

Another challenge of DL-based methods is that their performance usually decreases beyond the scope of training data. Notably, current AD-targeted datasets are inherently skewed toward individuals at high risk of developing AD, resulting in an imbalanced age distribution within the data (refer to Figure \ref{fig:datasetPartition}). To mitigate potential model bias toward the majority age group, we employ an oversampling strategy during training to get a similar representation of images at all ages during training. The outcomes illustrated in Figure \ref{fig:trends} show that the learned templates accurately capture underlying trends in different brain regions. As shown in that figure, the trend of the templates is approximately in the middle of the curve of inter-subject variability of the CN dataset. Moreover, these trends exhibit smoother trajectories compared to real data. Moreover, it can be noticed that the CN cohort has a lower inter-subject variability compared with the AD cohort. Although irrelevant to this study, this means that templates generated from the AD cohort might be less representative of the real data. 

Besides the imbalanced age distribution problem, we previously examined in \citet{fu2023fast} how domain shifts from training data to unseen data can damage registration performance, thereby requiring compensation to take place by introducing a stopping point different from one in the integration layer. Since the training and testing images in this study come from the same dataset, this adjustment is not required. Still, this is an issue to consider when multiple datasets are used.

\subsection{Aging and AD-specific scores}
Studies have demonstrated the intertwined relationship between aging and AD, suggesting that AD could partly reflect an accelerated brain aging process ~\cite{fjell2009one, whitwell2008rates, raz2005regional}. Moreover, it has been observed that normal aging manifests differently in different brain regions \cite{fjell2009one}, and certain structures such as the hippocampus and amygdala are more susceptible to AD \cite{xu2000usefulness, jack1999prediction, cavedo2011local}. However, precisely describing normal aging, particularly involving localized brain aging is complex. Furthermore, disentangling normal aging components from disease-specific changes poses additional challenges. Utilizing a refined score for normal aging at each voxel offers a means to observe disease-specific changes locally accurately.

We propose AS and ADS as features that could potentially be used as imaging biomarkers in AD. For this, we first define a unit-year normal aging SVF using DL-based diffeomorphic registration between the generated templates of CN subjects at 60 and 90 years old. This SVF captures the intricate transformations between templates, thereby identifying voxel-level anatomical changes associated with normal aging. In the next step, we register the given image to the 60-year-old template and perform a voxel-wise projection between the obtained SVF and the reference unit-year normal aging SVF. The parallel and orthogonal components of the projection are used to estimate AS and ADS, respectively. In order to increase the robustness of the estimations, voxels with too small unit-year normal aging are removed.
Finally, we applied a DL-based robust segmentation method to both the learned templates and the testing image in order to perform regional analyses.

\subsubsection{Evaluation on the CN cohort}
The performance of AS and ADS is evaluated using 326 MRI scans, all of which have a consistent CDR score of 0 across longitudinal MRI sessions. 
Numerous studies have highlighted the fact that brain aging occurs at varying rates across different brain regions \cite{raz2005regional, walhovd2005effects, bagarinao2022reserve}. While ventricle enlargement is a common observation in both the normal aging process and AD progression \cite{apostolova2012hippocampal, kwon2014age}, AD is characterized by more pronounced atrophy in regions such as the hippocampi and amygdala \cite{deture2019neuropathological, poulin2011amygdala, gosche2002hippocampal}. Employing segmentation allows us to perform regional analyses. In particular, we investigate ventricles, hippocampi, and amygdala regions in addition to the whole brain. By aligning different age cohorts to a common anatomy, such as the template of 60 years-old in our study, we enable meaningful inter-group structural comparisons.

As mentioned, we used a thresholding strategy to filter outliers based on the magnitude of $\bm v_0$ for the same anatomy. 
We examined the correlation between chronological age and the AS at different thresholding quantiles. We used $R^2$ to determine the best threshold per region as shown in Fig. \ref{fig:R2_num}. Considering an assumption that morphological age corresponds to chronological age, the AS should precisely reflect the normal aging process, exhibiting linear changes with age as shown in Figure \ref{fig:markers_on_HC} (assuming the reference template corresponds to the age of 60). However, morphological changes do not always exhibit a simple equivalent correlation with increasing age.
We find that different regions exhibit distinct optimal quantile thresholds for outlier removal. This variability might arise from factors such as considerable inter-subject anatomical variation.

Numerous neurodegeneration studies have highlighted the considerable inter-subject variation even within cognitively normal populations \cite{bagarinao2022reserve, walhovd2005effects, eavani2018heterogeneity, wrigglesworth2023health, ferreira2017cognitive}. This inherent variability also affects the estimations of the AS, which might contribute to the relatively low $R^2$ scores in Figure \ref{fig:markers_on_HC}. Although the $R^2$ scores for AS are relatively low, the p-values emphasize the significant correlation between chronological age and AS scores in the observed MRI regions. Conversely, despite the very small $R^2$ scores for ADS, indicating a weak correlation between chronological age and ADS scores, the p-values underscore the importance of accounting for age effects for certain MRI regions. Consequently, we employed a normalization procedure when comparing groups with substantial age distribution shifts.

Two notable sources of error deserve discussion. Firstly, the registration algorithm yields larger deformations for inter-subject registrations compared to intra-subject ones.
The intricate nature of brain sulci and giri, which might substantially vary among individuals, is analogous to a brain print similar to a fingerprint \cite{wachinger2015brainprint}. This makes it less evident the decomposition of the SVFs into their normal aging and AD-specific components in cortical regions. Secondly, we identified a potential registration error in regions that exhibit homogeneous intensity, such as the ventricles. In these cases, deformation vectors could become minute. For this reason, we perform regional analyses on the region close to the edges of the ventricles where registration is more reliable, as shown in the fourth column in Figs. \ref{fig:R2_num_markers_on_HC} and \ref{fig:markers_mag_on_AD}.

\subsubsection{Evaluation on the AD cohort}
Subsequently, the performance of AS and ADS are assessed in the AD cohort using the CN group as a comparison. The AD cohort comprised 688 MRI scans with varying CDR scores ranging from cognitively normal (CDR 0) to moderate AD dementia (CDR 2). It is worth mentioning that OASIS-3 focused on enrollment of early to mild AD individuals. That is why the CDR > 2 stages are not represented in the dataset. The terms “early stage” and “later stage” of AD were used in our discussion to indicate the relative stage in the range of early stage AD dementia of the dataset, which is more difficult to differentiate. Results are presented in boxplots along with p-value levels on each subfigure, as well as pair-wise Cohen's d estimates to quantify the effect sizes of the differences. 

When examining the ability of the two scores to tracking disease progression, we found that the two scores are indispensable and can be supplementary to each other. For example, by inspecting the boxplots, we found obvious ascending trends with an increase of disease stages in AD subgroups in all four observed brain regions for AS as illustrated in Figure \ref{fig:markers_on_AD}, and in hippocampi $\&$ amygdala regions for ADS in Figure \ref{fig:mag_on_AD}, despite the absence of a notable distinction across AD subgroups in chronological age distributions, as shown in Figure \ref{fig:Age_dist}.
In addition to the ascending trends over the whole AD course, the two scores exhibit different sensitivities in different stages of AD progression by examining the p-values and effect sizes. The ADS seems to have higher sensitivity for the later stage of AD. For example, there is no statistical difference by using AS in any MRI region for a later stage between CDR 0.5 vs CDR 1 pair until we consider the hippocampi $\&$ amygdala regions by using ADS for this pair. The bigger effect sizes for ADS compared with AS (i.e., -0.37 vs. -0.06 in the hippocampi $\&$ amygdala regions, -0.20 vs. -0.07 in the whole brain region in Table \ref{tab:effectsize}) in this pair further validates that. In contrast, AS has a better performance than ADS in the early stage, such as CDR 0 vs. CDR 0.5. The bigger effect sizes can be observed in all four MRI regions for AS compared with ADS (i.e., (0, 0.5) column in Table \ref{tab:effectsize}) in this pair. An intriguing observation is the noticeable differences between the CN and CDR = 0 AD subgroup by using both scores. This might suggest that morphological changes manifest prior to detectable cognitive alterations in clinical instruments such as the CDR.

Carrying out regional comparisons also reveals the different discriminative power of two scores among different MRI regions and extends the regional dimension to interpret two scores. For example, as shown in Figure \ref{fig:markers_on_AD}, except for the whole brain, all other three regions exhibit statistical differences during the progression of AD. These results support the hypothesis that AD accelerates aging in the ventricles and hippocampi \cite{fjell2009one, fox2004imaging, lorenzi2015disentangling}. In contrast, the ADS shows significant differences in the hippocampus and amygdala and no significant differences in ventricle-related regions within AD subgroups. The coupling of the results from the two scores suggests that ventricles predominantly follow accelerated normal aging in AD, while the atrophy in hippocampi $\&$ amygdala regions is influenced by both normal aging and AD-specific directions. By examining hippocampi $\&$ amygdala regions, which are affected by both normal aging and AD-specific related atrophy, we can observe from the quantified effect size in Table \ref{tab:effectsize} for those regions that a transferred atrophy pattern can be revealed by disentangling normal aging and AD-specific factors. For example, in the pair (CN, 0), the effect sizes of both scores are very similar (i.e., 0.22 vs. 0.24), while in the early stage in AD, AS contributes more than ADS (i.e., 0.41 vs. 0.32 in pair (0, 0.5)), then in the later stage, ADS dominates more than AS (i.e., 0.06 vs. 0.37 in pair (0.5, 1)). 

To summarize, the experiments on the OASIS-3 dataset using the two introduced scores revealed different atrophy patterns in terms of AD progression and regions affected by AD. Specifically, we show that the enlargement of ventricles in AD is contributed predominantly by accelerated normal aging direction and is less associated with morphological changes caused by AD-specific direction. In contrast, hippocampi $\&$ amygdala regions are influenced more by normal aging direction at early AD stages and then transferred to be influenced more by AD-specific direction at late AD stages. 

\subsection{Relationship with classical morphometric studies}
Voxel-based morphometry (VBM) and tensor-based morphometry (TBM) methods have been widely used in group analyses between cognitively healthy and AD groups \cite{scahill2002mapping, hua2008tensor, khan2015unified}. These methods usually employ scalars, such as the Jacobian determinant, to illustrate local expansion and contraction patterns in gray matter \cite{baron2001vivo}. While our proposed method also uses registration, the goal is to assess whether the detected changes in the brain follow the patterns of normal aging or not.

Our study adopts a deformation-based morphometric approach, enabling the isolation of the normal aging and AD-specific components—attributes not attainable through VBM or TBM.
Central to conventional morphometric group-wise studies is the construction of a target-specific template. Historically, this process involved resource-intensive registration computations between each pair of intra-subject imaging data, occasionally necessitating the use of mathematically demanding tools like parallel transport to transfer individual trajectories to template space \cite{hadj2016longitudinal, lorenzi2015disentangling}. However, in our study, we take advantage of the efficiency of deep learning methods for both template creation and registration steps.
By leveraging the power of deep learning, our framework offers several advantages over traditional morphometric approaches. The integration of DL-based diffeomorphic registration not only enhances the accuracy of template creation but also facilitates the extraction of nuanced anatomical patterns associated with normal aging in an efficient manner.  

\subsection{Limitations and future work}
Some limitations should be acknowledged. Firstly, our framework is trained and evaluated solely on the OASIS-3 dataset, which may introduce dataset-specific biases and potentially limit the generalizability of our findings. For example, the creation of normal aging templates is based on 1352 scans from the CN group within this dataset, which might be biased by the fact that OASIS-3  predominantly included participants in specific regions (i.e., primarily the United States) and exhibits an unbalanced representation of racial diversity (e.g., only 5 Asian participants are reported in the demographics). Consequently, the normal aging patterns derived from this dataset might not be fully representative of the broader human population. So one needs to consider this effect when applying the learned normal aging pattern to datasets from different domains. Although other publicly available datasets, e.g., ADNI, are extensively used for AD research, we decided not to incorporate it in this study to avoid issues associated with domain shifts between different datasets \cite{fu2023fast}, ensuring more reliable and interpretable results as assessed in the same dataset.

Additionally, it is important to acknowledge that the approach of determining the AD-specific component through the average magnitude of the residual vector is not without limitations. The registration process between a reference CN template and each subject's scan can be considered an inter-subject scenario. Consequently, the residual vector may contain both AD-specific and subject-specific components. While steps were taken to mitigate the influence of subject-specific differences through affine registration and an outlier rejection strategy, the development of a more sophisticated model that explicitly addresses these factors is a worthwhile pursuit for achieving a more accurate and interpretable model. Some methods have emerged to tackle this inter-subject variability by explicitly modeling aging and disease severity. For instance, \citet{ouyang2022disentangling} introduced orthogonal constraints within the latent space of a variational autoencoder. However, this approach's learned global coefficient does not offer regional information. Another approach, as seen in \citet{sivera2019model}, explicitly models both components along with an inter-subject component through longitudinal data. While the incorporation of longitudinal data offers potential benefits in building a comprehensive and robust model, such data is often limited within a single dataset. Future studies could consider augmenting the datasets with more longitudinal data or including a bigger longitudinal dataset to better address intra-subject differences. It is worth noting that while our approach does not specifically account for subject differences, the results remain valid when comparing across groups under the assumption that group-wise subject-level differences are consistent.

\section{Conclusion}
In summary, our framework can be used to investigate the intricate interplay between AD and normal aging in terms of brain atrophy, addressing critical questions about their relationship. Leveraging deep learning advancements, we construct age-specific templates and delineate normal aging atrophy patterns through advanced registration techniques. The extracted voxel-level vectors reveal nuanced variations attributed to normal aging, fostering a more robust evaluation of their connection to AD progression. Our study uncovers that the impact of AD is twofold: the trajectory of normal aging-related brain atrophy gets faster with AD in certain brain regions (e.g., both ventricles and hippocampi \& amygdala), while other regions exhibit distinctive AD-specific atrophy patterns (e.g., hippocampi \& amygdala). Combining these components may in the future enable differentiation of subtle AD clinical stages. In essence, our work contributes valuable insights into the convergence of normal aging and AD.

\section*{Funding}  
This study has been partially funded by the Swedish Childhood Cancer Foundation (Barncancerfonden; MT2019-0019, MT2022-0008), by Vinnova through AIDA, project ID: 2108, by the China Scholarship Council (CSC) for PhD studies at KTH Royal Institute of Technology, by Digital Futures, project dBrain, by the Swedish Research Council (Vetenskapsrådet, grant 2022-00916), the Center for Innovative Medicina (CIMED, grants 20200505 and FoUI-988826), the regional agreement on medical training and clinical research of Stockholm Region (ALF Medicine, grants FoUI-962240 and FoUI-987534), the Swedish Brain Foundation (Hjärnfonden FO2023-0261, FO2022-0175, FO2021-0131), the Swedish Alzheimer Foundation (Alzheimerfonden AF-968032, AF-980580), the Swedish Dementia Foundation (Demensfonden), the Gamla Tjänarinnor Foundation, the Gun och Bertil Stohnes Foundation, Funding for Research from Karolinska Institutet, Neurofonden, and Foundation for Geriatric Diseases at Karolinska Institutet. The funders of the study had no role in the study design nor the collection, analysis, and interpretation of data, writing of the report, or decision to submit the manuscript for publication.
 
\section*{Acknowledgements}  

Data were provided in part by OASIS-3: Principal Investigators: T. Benzinger, D. Marcus, J. Morris; NIH P50 AG00561, P30 NS09857781, P01 AG026276, P01 AG003991, R01 AG043434, UL1 TR000448, R01 EB009352. AV-45 doses were provided by Avid Radiopharmaceuticals, a wholly owned subsidiary of Eli Lilly.  

\section*{Author contributions}
\textbf{JF}: conceptualization; formal analysis; investigation; methodology; software; validation; visualization; writing - original draft; writing - review \& editing. 
\textbf{DF}: methodology; writing - review \& editing; supervision.
\textbf{ÖS}: resources; writing - review \& editing; supervision; funding acquisition.
\textbf{RS}: conceptualization; methodology; visualization; resources; project administration; writing - review \& editing; supervision; funding acquisition. 

\section*{Conflict of Interest}
The authors declare no conflict of interest.

\section*{Data Availability Statement}
The data that support the findings of this study are openly available in OASIS at
\url{http://doi.org/10.1101/2019.12.13.19014902}, reference number~\citep{lamontagne2019oasi}.
The source codes generated for this study are available on \url{https://github.com/Fjr9516/DBM_with_DL}.

\bibliographystyle{unsrtnat}
\bibliography{references}  






\end{document}